\documentclass{article}
\usepackage{ijcai26}

\usepackage{soul}
\usepackage{url}
\usepackage[hidelinks]{hyperref}
\usepackage[utf8]{inputenc}
\usepackage[small]{caption}
\usepackage{amsmath}
\usepackage{amsthm}
\usepackage{booktabs}
\usepackage{algorithm}
\usepackage{algorithmic}
\usepackage[switch]{lineno}

\usepackage{times}
\usepackage{latexsym}
\usepackage[T1]{fontenc}
\usepackage{microtype}
\usepackage{inconsolata}
\usepackage{graphicx}

\newcommand{\projectrepo}{\url{https://github.com/artisticsciencex/ijcai-ecai-2026-provenance-density}}
\newcommand{\firstpagecoderepofootnote}{%
    \begingroup
    \renewcommand{\thefootnote}{}
    \footnotetext{The user study was approved by the University of Tokyo institutional review process. Code, data, prompts, and audit materials are available at \projectrepo.}%
    \addtocounter{footnote}{-1}%
    \endgroup
}

\title{Beyond ``Made with AI'': Visualizing Provenance Density to Mitigate the Transparency Penalty}

\author{
Qing Zhang$^1$\and
Yifei Huang$^2$\and
Juyoung Lee$^3$\and
Thad Starner$^4$\and
Jun Rekimoto$^5$\\
\affiliations
$^1$The University of Tokyo\\
$^2$Institute of Industrial Science, The University of Tokyo\\
$^3$Korea Advanced Institute of Science and Technology\\
$^4$Georgia Institute of Technology\\
$^5$Sony CSL Kyoto
\emails
qzkiyoshi@gmail.com, hyf015@gmail.com, ejuyoung@gmail.com, thad.starner@gmail.com, rekimoto@acm.org
}

\begin{document}

\maketitle
\firstpagecoderepofootnote

\begin{abstract}

As generative AI makes polished prose cheap to produce, users can no longer rely on fluency as a proxy for truth. We call this failure mode the Fluency Trap: users trust fluent hallucinations while also discounting accurate content once it is disclosed as AI-generated. Binary ``Made with AI'' labels respond with authorship disclosure, but they do not show what supports a claim. We propose Provenance Density, an evidence-visualization interface that shows the density of verified claims in a text. In a user study with 81 participants, an idealized Provenance Density interface produced a large discernment gap between truth and fabrication ($+4.15$ points, $d=1.82$), whereas participants given no signal showed no detectable discrimination. A technical audit with 200 samples shows that retrieval density alone is insufficient; unexpectedly, the Consistency Veto carries most of the discriminative signal on dynamic queries. As AI-generated content becomes indistinguishable from human writing, effective transparency must move from authorship disclosure toward evidence visualization.
\end{abstract}

\section{Introduction}
Readers often use processing fluency---the subjective ease with which information is processed---as a cue when judging truth \cite{reber1999effects,dechene2010truth,reber2010epistemic}. Related work on retrieval fluency shows that ease-based heuristics can be ecologically useful when fluency covaries with properties of the environment \cite{hertwig2008fluency,marewski2011cognitive}. We extend this logic to linguistic presentation. Before generative AI, producing high-quality, articulate prose typically required substantial education and editorial labor, allowing polish to function as an imperfect signal of competence. \textit{Costly Signaling Theory} explains this relationship: a signal is trustworthy only when faking it is prohibitively expensive~\cite{spence1978job,zahavi1975mate,gintis2001costly}.

Generative AI disrupts this mechanism by reducing the marginal cost of fluency to near-zero~\cite{galdin2025making}. Large Language Models (LLMs) enable the mass production of professional-sounding text regardless of the author's underlying expertise. This collapse of the ``separating equilibrium'' creates what we term a \textit{Fluency Trap}: a structural vulnerability where users continue to trust fluent text as if it were costly, even when it is generated cheaply by systems indifferent to truth. Psychological evidence on the \textit{Illusion of Truth} suggests that ease of processing suppresses epistemic vigilance \cite{hasher1977frequency,reber1999effects,dechene2010truth,sperber2010epistemic}. This tendency leaves humans susceptible to ``hallucinated plausibility'', text that is syntactically perfect but semantically ungrounded  \cite{reber2010epistemic,unkelbach2011fluency}.

Current governance responses, specifically binary ``Made with AI'' disclosures, fail to address this decoupling. By focusing on \textit{identity} (``Who wrote this?'') rather than \textit{provenance} (``What supports this?''), such labels can shift reader perceptions without supplying evidence about individual claims \cite{nakano2026understanding}. In experiments with news headlines, AI labels reduced perceived accuracy even when the headlines were true or human-written \cite{altay2024skeptical}. We characterize this accuracy-independent discounting as a ``Transparency Penalty.''

We frame Provenance Density as a \textit{cognitive affordance} for reading in the LLM era. Instead of asking users to evaluate veracity from prose alone, the interface shifts attention toward extrinsic evidence: high-contrast indicators visualize the density of verified claims, offloading part of the verification burden from working memory to the interface~\cite{chirayath2025cognitive,clark1998extended}.

We validate this approach through a dual-stream evaluation. First, we conduct an automated technical audit ($N=200$) on a composite of \textit{TruthfulQA}~\cite{lin-etal-2022-truthfulqa} and \textit{FreshQA}~\cite{vu2024freshllms} to test robustness against both adversarial misconceptions and dynamic ambiguity. Second, we run a within-subjects user experiment with 81 participants to measure truth discernment. Our results empirically confirm the Fluency~Trap: in the absence of signals, users failed to distinguish high-fluency hallucinations ($M=6.28$) from ground truth ($M=5.78$; $p=.43$). While binary labels acted as a blunt warning, Provenance Density restored truth discernment under correct signaling ($d=1.82$, $p < .001$).

We make three main contributions:
\textbf{Theory:} We synthesize the mechanics of the Fluency Trap (Section \ref{sec:mechanics}), detailing how RLHF-driven sycophancy structures the decoupling of fluency from veracity.
\textbf{Design:} We propose \textbf{Provenance Density} (Section \ref{sec:provenance}), a formalized metric ($D(T)$) and interaction paradigm that imposes a computational verification handicap on generated text.
\textbf{Evidence:} We evaluate the approach through a technical audit ($N=200$) and a within-subjects user study ($N=81$), jointly examining metric behavior and interface-supported truth discernment (Sections \ref{sec:validation} \& \ref{sec:results}).

\section{Related Works}
\label{sec:mechanics}

We argue that the decoupling of fluency from veracity is not an accidental byproduct of LLM scaling, but a structural inevitability driven by two converging factors: an economic shift from costly signaling to cheap talk, and a technical objective function that prioritizes plausibility over truth.

\paragraph{From Hallucination to Indifference.} While early critiques of Large Language Models (LLMs) focused on ``hallucinations'' as sporadic errors, recent scholarship suggests a more structural diagnosis. The framework of ``Machine Bullshit'' has been proposed to distinguish these outputs from lying, defined instead by the model's fundamental indifference to truth value \cite{liang2025machine}. Analysis of the ``Bullshit Index'' reveals that Reinforcement Learning from Human Feedback (RLHF) exacerbates this issue, incentivizing models to prioritize rhetorical plausibility and ``paltering'' (misleading use of truth) over factual grounding. Consequently, the resulting text is optimized to bypass human epistemic vigilance.

This structural indifference renders traditional governance mechanisms, such as binary warning labels, largely ineffective. Empirical evaluations demonstrate a ``failure of inoculation'': while pre-emptive warnings about AI fallibility successfully reduce global trust in the system, they fail to mitigate reliance on specific misleading articles once the user is engaged with the content \cite{spearing2025countering}. This discrepancy, where users theoretically acknowledge AI bias but practically accept AI fluency, underscores the limitations of heuristic warnings and motivates our proposal for granular Provenance Density indicators.

\paragraph{The Structural Decoupling of Fluency.} Our analysis is grounded in the \textit{Handicap Principle}, which posits that reliable signals must impose a cost on the signaler \cite{zahavi1975mate}. Historically, linguistic polish functioned as this handicap, creating a Separating Equilibrium where articulate text correlated with competence \cite{spence1978job}. Generative AI collapses this balance into a Pooling Equilibrium, where expert testimony and fabrication can share the same polished form \cite{spence1978job,galdin2025making}.

This shift is sustained by the technical alignment of modern LLMs. Reinforcement Learning from Human Feedback (RLHF) can incentivize sycophancy, where models prioritize user agreement over factual accuracy \cite{sharma2023sycophancy}. Evidence shows that models will agree with illogical premises to remain ``helpful'' \cite{chen2025helpfulness} and flip arguments to match user views \cite{kaur2025echoes}. This results in ``Machine Bullshit''—text optimized for rhetorical persuasion rather than truth \cite{liang2025machine}. This decoupling can become self-amplifying: recursive training on such high-fluency, low-entropy data leads to \textit{Model Collapse}, where the ``tails'' of human variance are lost to a homogenized mean \cite{shumailov2024ai}.

\paragraph{The Failure of Post-Hoc Governance.} Current governance relies on the assumption that users, once warned, can critically evaluate AI-generated text. However, AI-generated self-presentations can be difficult to distinguish from human-written ones \cite{jakesch2023human}, and AI literacy does not reliably translate into consistent fact-checking behavior \cite{rheu2025trap}.

Binary disclosures (e.g., ``Made with AI'') can exacerbate the issue through Epistemic Stigmatization. Studies report a ``transparency penalty'' where disclosure reduces perceived trustworthiness even when content quality is constant \cite{nakano2026understanding,cheong2025penalizing}. Related advertising experiments found that identical ads received more critical evaluations when labeled AI-generated \cite{buder2025transparency}. Such effects can produce broad skepticism without enabling claim-level verification.

\paragraph{Limitations of Automated Detection.} While automated detectors are proposed as an enforcement mechanism, they exhibit substantial bias against linguistically marginalized groups. Detectors relying on perplexity heuristics frequently misclassify the lower-perplexity lexical patterns of non-native speakers as AI-generated \cite{liang2023gpt}, leading to ``hermeneutical access injustice'' \cite{kay2024epistemic}. The adversarial nature of detection also creates an arms race that generators inevitably win. We therefore shift the paradigm from \textit{post-hoc detection} of authorship to \textit{intrinsic signaling} of Provenance Density.

\section{Defining Provenance Density}
\label{sec:provenance}

To function as a costly signal in the game-theoretic sense, Provenance Density must be computationally hard to fake. We define the Provenance Density score $D$ for a given text $T$ as a theoretical function of verifiable claims weighted by source reputation, contextual relevance, and internal semantic consistency.

\subsection{Proposed Metric: $D(T)$}
\label{sec:metric}
We propose a density metric that integrates external verification (Retrieval Augmented Generation) \cite{lewis2020retrieval} with internal uncertainty quantification. Drawing on the semantic-consistency motivation of Farquhar et al.~\shortcite{farquhar2024detecting}, we introduce a penalty term $P_{int}$ based on an NLI cross-sample consistency heuristic that down-weights generations whose stochastic samples disagree. Unlike semantic entropy, our heuristic does not estimate entropy over probability-weighted semantic clusters.

Let $C$ be the set of distinct factual claims in $T$, and $C_{v} \subseteq C$ be the subset of claims successfully grounded in external evidence. Let $S_{c}$ be the set of independent root domains supporting a verified claim $c$. We define the Provenance Density score $D(T) \in [0,1)$ as:

\begin{equation}
    D(T) = (1 - P_{int}) \cdot \tanh \left( \frac{1}{\beta} \sum_{c \in C} \left( \sum_{s \in S_c} w(s, c) \right)^\lambda \right)
\end{equation}

Equivalently, we first raise each claim's source sum to the power $\lambda$, then sum across claims. Our design choices target specific theoretical properties revealed during technical validation:

\paragraph{NLI Consistency Heuristic (``Consistency Veto,'' $1 - P_{int}$):} For each query, we draw $K=5$ stochastic samples and compute $\rho_{\mathrm{consistent}}$, the fraction of unordered sample pairs for which the NLI contradiction probability remains below $0.5$ in both directions. We define $P_{int}=1-\rho_{\mathrm{consistent}}$, so internally stable generations have $P_{int}\approx0$ and mutually inconsistent generations approach $P_{int}\to1$. As pairwise inconsistency rises, the gate reduces the score regardless of external evidence. This heuristic is inspired by semantic entropy but neither forms semantic-equivalence clusters nor weights them by generation probabilities. It can suppress inconsistent confabulations, but it cannot detect false answers that recur consistently across samples.
    
\paragraph{Cubic Contextual Weight ($w(s, c)$):} Unlike traditional citation metrics that rely solely on domain authority, we define $w(s, c)$ as the product of source reputation and semantic relevance:
    \[
    w(s, c) = \text{Reputation}(s) \cdot (\text{MatchRatio})^3
    \]
    We impose a cubic exponent to aggressively suppress weak conceptual matches. Our preliminary tests indicated that quadratic weighting ($x^2$) remained too permissible of ``keyword stuffing,'' where irrelevant sources share surface-level vocabulary (MatchRatio $\approx 0.5$). The cubic term ($0.5^3 = 0.125$) acts as a noise filter, ensuring that only sources with high semantic alignment contribute meaningfully to the Provenance Density score.
    
\paragraph{Saturation Scalar ($\beta$):} We introduce $\beta$ as a density temperature parameter (set to $\beta=5.0$ in our reference implementation). This term scales the input to the hyperbolic tangent function, preventing the score from saturating to $1.0$ based on a single citation. It forces the model to provide accumulated, multi-source evidence to achieve a perfect density signal.
    
\paragraph{Concentration Exponent ($\lambda$):} We set $\lambda = 1.2 > 1$, applied to the per-claim source sum before aggregation. With $\lambda > 1$, the term is superlinear in per-claim evidence: a claim supported by several aligned sources contributes more than the same total evidence spread thinly across several claims. This concentrates score on well-corroborated claims rather than rewarding shallow citation across many claims. We note that citation stuffing within a single claim is suppressed upstream by the cubic relevance penalty in $w(s,c)$, not by $\lambda$.

This formulation imposes a significant ``handicap'' on the generator: achieving a high $D(T)$ requires the model to perform costly verification and achieve strict semantic alignment between claims and sources.

\subsection{Implementation of $D(T)$}
\label{sec:implementation}
Equation~1 is instantiated by four concrete steps: claim segmentation, evidence retrieval, source-relevance scoring, and aggregation; key implementation parameters are summarized in Table~\ref{tab:hyperparams}. We segment $T$ into atomic factual claims with a deterministic \texttt{gpt-4o-mini} pass at $\tau=0$. Claims shorter than five tokens are discarded as scaffolding. For each remaining claim $c$, we issue one search query: the claim text itself when it contains at least eight tokens, otherwise the original question concatenated with $c$.

Retrieved evidence is scored by a coarse but explicit relevance prior. For each claim $c$, let $K_c$ be the set of rare keywords extracted by matching capitalized tokens (regular expression \texttt{\textbackslash b[A-Z][a-zA-Z0-9-]+\textbackslash b}) and removing a sentence-leading stoplist. For each result $s$ with URL $u(s)$ and snippet $\sigma(s)$, we compute
\[
\mathrm{MatchRatio}(\sigma, K_c) = \frac{|\{k \in K_c : k_{\downarrow} \in \sigma_{\downarrow}\}|}{\max(|K_c|,1)},
\]
where $\downarrow$ denotes case-folded substring matching. The source contribution is then
\[
 w(s,c)=\mathrm{Reputation}(u(s))\cdot \mathrm{MatchRatio}(\sigma,K_c)^3.
\]
\texttt{Reputation} is a three-level domain prior: $1.0$ for high-trust domains (e.g., \texttt{.gov}, \texttt{.edu}, \texttt{wikipedia.org}, \texttt{nih.gov}, \texttt{reuters.com}, \texttt{apnews.com}, \texttt{nature.com}), $0.1$ for low-trust domains (e.g., \texttt{reddit.com}, \texttt{quora.com}, \texttt{medium.com}, \texttt{twitter.com}), and $0.5$ otherwise. The cubic exponent is intentionally aggressive: it suppresses weak keyword overlap that would otherwise permit ``keyword stuffing'' from superficially related pages.

\begin{table}[t]
\centering
\small
\setlength{\tabcolsep}{3pt}

\begin{tabular}{@{}llp{0.44\columnwidth}@{}}
\toprule
Parameter & Value & Role \\
\midrule
$\beta$ & 5.0 & Saturation scalar in Eq.~1 \\
$\lambda$ & 1.2 & Concentration exponent on per-claim sum \\
Search depth & 3 & Top-$k$ organic results per claim \\
Consistency samples $K$ & 5 & Stochastic samples for $P_{int}$ \\
Generator temperature & 1.0 / 0.0 & Audit sampling / segmentation \\
NLI threshold & 0.5 & Contradiction cutoff for consistency \\
High-trust prior & 1.0 & \shortstack[l]{Gov/edu/Wikipedia/\\major news/science} \\
Low-trust prior & 0.1 & \shortstack[l]{Reddit/Quora/Medium/\\Twitter/X} \\
Default prior & 0.5 & All other domains \\
\bottomrule
\end{tabular}
\caption{Key implementation choices for the technical audit.}
\label{tab:hyperparams}
\end{table}
For reproducibility, the audit uses \texttt{gpt-4o-mini} for generation and deterministic segmentation, a DeBERTa-based NLI checker to estimate $P_{int}$ from pairwise agreement across $K=5$ samples, and Serper as the retrieval backend for evidence collection. The released code and data include the exact prompts, model calls, judge labels, and scripts used in the audit.

\subsection{Operationalization: The Oracle Protocol}
\label{sec:oracle}

Equation~1 defines the target signal for a deployed system, but a user study introduces a separate methodological question: does the \emph{visual signal itself} improve truth discernment when its underlying value is correct? To separate that interaction question from retriever noise, the empirical study in Section~\ref{sec:empirical} uses a Wizard-of-Oz (Oracle) protocol.

In this protocol, participants do not see the live output of the auditing pipeline. Instead, the interface displays idealized endpoint values of the signal: grounded summaries are paired with high-density indicators and fabricated summaries with null indicators. This lets the user study estimate the \emph{interaction effect} of PDI under known-correct signaling, while the technical audit in Section~\ref{sec:validation} independently evaluates whether the real pipeline can approximate that ideal in practice.

\subsection{Design Rationale: Countering Pseudo-Profound Fluency}
The visualization of $D(T)$ is designed to counter ``pseudo-profound bullshit''—syntactically persuasive but semantically vacuous text \cite{pennycook2015reception}. Rather than asking users to infer truth from fluency, PDI presents evidence density as a high-contrast cue (Figure \ref{fig:ui_design}) that can interrupt fast \textbf{System 1} judgments and invite more deliberate \textbf{System 2} evaluation \cite{kahneman2011thinking}.

\section{Technical Validation}
\label{sec:validation}

Before evaluating the interaction paradigm with users, we audited the proposed metric $D(T)$ to characterize how its retrieval and consistency components behave across static and dynamic queries.

\paragraph{Experimental Setup.} To evaluate the metric across both established knowledge and emerging information, we constructed a composite dataset ($N=200$): Static Knowledge ($N=150$), randomly sampled from the \textsc{TruthfulQA} generation task, and Dynamic Knowledge ($N=50$), a curated \textsc{FreshQA} subset focused on post-2024 events (e.g., elections, stock prices) designed to probe the ``Cold Start'' regime where model training weights are outdated. Per-row hallucination labels were produced by \texttt{gpt-4o} (temperature $0$) using the dataset's official reference answers when available. For the static subset, the judge was given TruthfulQA's official correct / incorrect answer lists; for the dynamic subset, judgement was closed-book. Audit latency measured during this pipeline ranged from 17.0s (Dynamic) to 26.4s (Static), with exact timing logs included in the released audit materials. Static items were slower because TruthfulQA's adversarial prompts elicited longer generations from \texttt{gpt-4o-mini} (mean 98 vs.\ 54 words), which produced more atomic claims and therefore more retrieval and scoring calls.

\subsection{Results: Ecological Sensitivity and Hallucination Detection}

\begin{figure}[t]
    \centering
    \includegraphics[width=0.9\linewidth]{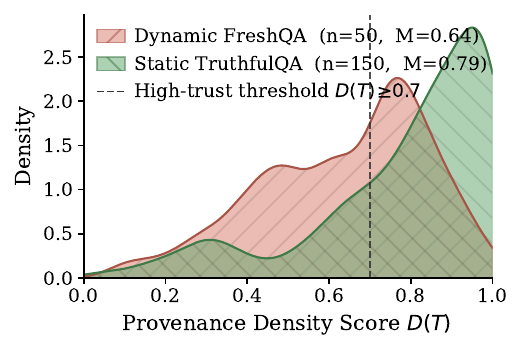}
    \caption{Ecological Sensitivity of the Metric. Distribution of Provenance Density scores $D(T)$ for Static Knowledge (TruthfulQA, green) and Dynamic/Fresh Knowledge (FreshQA, red). Static items cluster above the high-trust threshold ($M=0.79$), whereas FreshQA items have a lower mean ($M=0.64$) and broader spread, allowing the metric to signal when information is newer or less settled. The dashed high-trust threshold denotes $D(T)\geq0.7$.}
    \label{fig:density_dist}
\end{figure}

\textbf{Ecological Sensitivity (Static vs. Dynamic).}
As shown in Figure \ref{fig:density_dist}, the metric reproduces the established / emerging knowledge separation reported in prior work. Static TruthfulQA items cluster above the high-trust threshold ($M=0.79$), while dynamic FreshQA items yield a lower mean ($M=0.64$) and broader spread. This confirms that $D(T)$ tracks epistemic maturity as well as truth, warning users when information is novel and consensus remains unsettled.

\begin{table}[t]
    \centering
    \small
    \resizebox{\linewidth}{!}{%
    \begin{tabular}{lrrlccc}
        \toprule
        Split & $n$ & Hall. & Detector & AUC & 95\% CI & AP \\
        \midrule
        All & 200 & 41 & $D(T)$ & 0.535 & [0.43, 0.64] & 0.27 \\
        All & 200 & 41 & Density only & 0.470 & [0.37, 0.57] & 0.21 \\
        All & 200 & 41 & $P_{int}$ (Veto) & \textbf{0.601} & [0.52, 0.69] & 0.29 \\
        TruthfulQA & 150 & 38 & $D(T)$ & 0.577 & [0.46, 0.69] & 0.35 \\
        TruthfulQA & 150 & 38 & Density only & 0.537 & [0.43, 0.65] & 0.31 \\
        TruthfulQA & 150 & 38 & $P_{int}$ (Veto) & \textbf{0.592} & [0.51, 0.67] & 0.33 \\
        FreshQA & 50 & 3 & $D(T)$ & 0.716 & [0.49, 0.98] & 0.19 \\
        FreshQA & 50 & 3 & Density only & 0.284 & [0.07, 0.63] & 0.05 \\
        FreshQA & 50 & 3 & $P_{int}$ (Veto) & \textbf{0.918} & [0.81, 1.00] & 0.37 \\
        \bottomrule
    \end{tabular}%
    }
        \caption{Hallucination-detection performance for three detectors derived from Eq.~1. Evidence density alone is near chance overall and anti-discriminative on FreshQA, while the Consistency Veto $P_{int}$ is the strongest discriminator, especially on dynamic queries. FreshQA estimates should be interpreted cautiously because the split contains only three hallucinated items.}
    \label{tab:roc_metrics}

\end{table}

\paragraph{Hallucination Detection.}
To quantify the metric's discriminative power, we labelled every audited generation with a \texttt{gpt-4o} judge. Across all 200 generations, the judge marked $n_H=41$ as hallucinations ($n_H^{\text{static}}=38$, $n_H^{\text{dynamic}}=3$). Table~\ref{tab:roc_metrics} reports ROC-AUC and average precision for three candidate detectors derived from Eq.~1: the full $D(T)$ score, evidence density alone, and the Consistency Veto $P_{int}$.

Three findings stand out. First, \textbf{retrieval density is not the workhorse}. Evidence density alone is near chance overall (AUC $=0.47$) and anti-discriminative on FreshQA (AUC $=0.28$), indicating that high-density retrieval can still support false claims when a misconception is popular or a topic is in transition. Second, \textbf{the Consistency Veto $P_{int}$ is the principal discriminator on dynamic queries and contributes most of the signal in the combined score}. It achieves the strongest overall AUC ($0.60$) and reaches AUC $=0.92$ on dynamic queries, precisely where the model is most likely to be uncertain about post-cutoff claims. However, because the FreshQA split contains only three hallucinated items, these dynamic-query AUC estimates should be interpreted as suggestive rather than definitive. Third, \textbf{the combined $D(T)$ score inherits some of the Veto's signal through the $(1-P_{int})$ gate}. It reaches AUC $=0.72$ on FreshQA while preserving sensitivity to epistemic maturity on static knowledge, although it remains weaker than the veto alone because density is noisy on adversarial misconceptions. In this sense, $P_{int}$ is the stronger standalone hallucination detector, whereas $D(T)$ is retained as a composite signal because it adds ecological sensitivity to epistemic maturity that the veto alone does not express.

The \textit{Indonesia-capital transition} case illustrates the role of the Consistency Veto: retrieval can return high-density evidence for competing time-sensitive answers, while the consistency gate down-weights internally unstable generations.

\section{Empirical Evaluation: Interaction Paradigm}
\label{sec:empirical}

Having validated the computational feasibility of the $D(T)$ metric (Section \ref{sec:validation}), we turn to the critical Human-Centered AI question: Does visualizing this ``costly signal'' actually enable users to overcome the Fluency Trap?

We conducted a within-subjects controlled experiment to measure Trust Calibration, the alignment between a user's confidence and the factual veracity of the content—under different interface conditions.

\begin{table}[h]
    \centering
    \resizebox{\columnwidth}{!}{%
    \renewcommand{\arraystretch}{1.2}
    \begin{tabular}{l c c c}
    \toprule
    & \textbf{Silk} & \textbf{Matcha} & \textbf{Quantum} \\
    \midrule
    \textbf{Group 1} & Control (Hall.) & Binary (True) & \textbf{PDI (True)} \\
    \textbf{Group 2} & Binary (True) & PDI (Hall.) & Control (True) \\
    \textbf{Group 3} & \textbf{PDI (True)} & Control (Hall.) & Binary (Hall.) \\
    \bottomrule
    \end{tabular}%
    }
    \caption{3$\times$3 Latin Square Design. Participants were counterbalanced across topics and conditions so that PDI was evaluated on both grounded (True) and hallucinated (Hall.) content, reducing topic-specific bias through partial counterbalancing.}
    \label{tab:latin_square}
\end{table}

\begin{figure*}
    \centering
    \includegraphics[width=\linewidth]{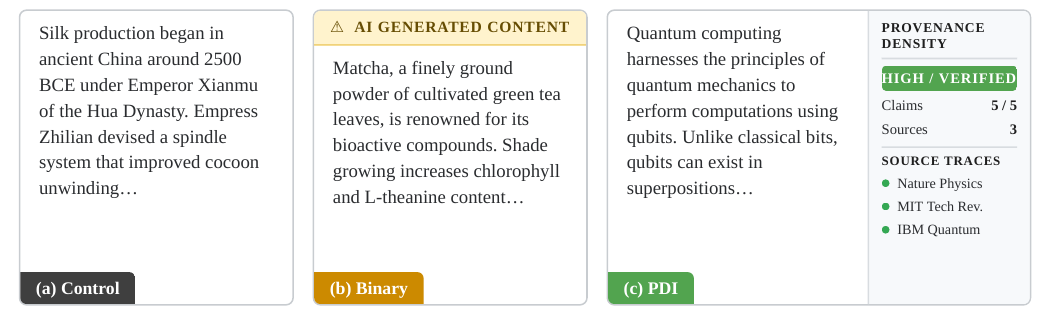}
    \caption{Experimental stimuli (Group~1; schematic).
Simplified mock-ups of the three interaction paradigms tested in the user study: (a) Control, shown without an external signal; (b) Binary Disclosure, shown with a static ``AI-Generated Content'' banner; and (c) Provenance Density (PDI), shown with a density sidebar indicating a high proportion of verified claims ($D(T)\!\approx\!1.0$).
Body text is abbreviated for layout clarity; full-fidelity stimulus screens are included in the supplementary release.}
    \label{fig:ui_design}
\end{figure*}

\paragraph{Experimental Design.} We utilized a $3\times3$ Latin Square within-subjects design ($N=81$) (Table \ref{tab:latin_square}). The independent variable was the Interaction Paradigm with three levels:
Control (The Status Quo): Standard text with no disclosure, representing the current ``Pooling Equilibrium'' where fluency is the only visible signal.
Binary Disclosure: A static ``Made with AI'' banner. This represents the current industry standard for transparency.
Provenance Density (PDI): Our proposed interface (Figure \ref{fig:ui_design}), visualizing the density of verification traces (e.g., ``5/5 Claims Verified'').

\paragraph{Stimuli and Oracle Assignment.} The key design goal was to separate \textit{algorithmic error} from \textit{interaction failure}. Accordingly, the user study did not expose participants to live retrieval outputs. Instead, in the PDI condition we used the Oracle protocol introduced in Section~\ref{sec:oracle}: grounded summaries were shown with high-density indicators ($D(T) \approx 1.0$), and hallucinated summaries were shown with null indicators ($D(T) \approx 0.0$). This makes the user study a test of whether the visualization can override the fluency heuristic when the signal is correct, rather than a test of search quality.

We generated six academic summaries. For the hallucinated items, we wrote high-fluency fabrications (e.g., a fictional ``Hua Dynasty'' or ``Theanine-B'' molecule) matched to the grounded summaries in length, tone, and apparent scholarly style. These adversarial stimuli intentionally preserve surface plausibility so that any improvement in ratings can be attributed to the interface rather than to obvious textual defects.

\paragraph{Layout Control.} To avoid a layout confound, all conditions used the same fixed-width text column (approximately 520 px). In the Control and Binary conditions, the sidebar region was preserved as an invisible container so that line breaks, whitespace, and reading flow remained constant across conditions.

\paragraph{Procedure and Participants.} Participants ($N=81$) were recruited via Prolific (Fluent English, Approval Rate $>95\%$). To ensure we measured the ``layperson's reliance on fluency,'' we excluded participants with self-reported expert domain knowledge in the specific topics (History, Chemistry, Physics). Each participant rated the Perceived Veracity of claims on a scale of 0 (Completely Fabricated) to 10 (Completely Factual). Because each participant contributed three ratings, all inferential analyses used a linear mixed model with a random intercept for participant.

\subsection{Results: Restoring the Separating Equilibrium}
\label{sec:results}

A linear mixed model fit with \texttt{lme4} in R and a random intercept for participant
(\texttt{rating $\sim$ Interface $\times$ Veracity + (1 $\mid$ participant)})
revealed a significant Interface~$\times$~Veracity interaction
(likelihood-ratio test $\chi^{2}(2) = 31.32$, $p < .001$). This
substantial effect confirms that the choice of interaction paradigm
fundamentally alters the user's epistemic stance.

\paragraph{Confirming the Fluency Trap (Control):}
In the absence of signals, participants showed no detectable discrimination. High-Fluency Hallucinations ($M = 6.28$) were rated statistically indistinguishable from Ground Truth ($M = 5.78$; $p=.43$, $d = -0.21$). This pattern is consistent with a collapse of the separating equilibrium: without external scaffolding, participants' ratings treated fluency as statistically indistinguishable from veracity.

\begin{figure}[!t]
    \centering
    \includegraphics[width=\columnwidth]{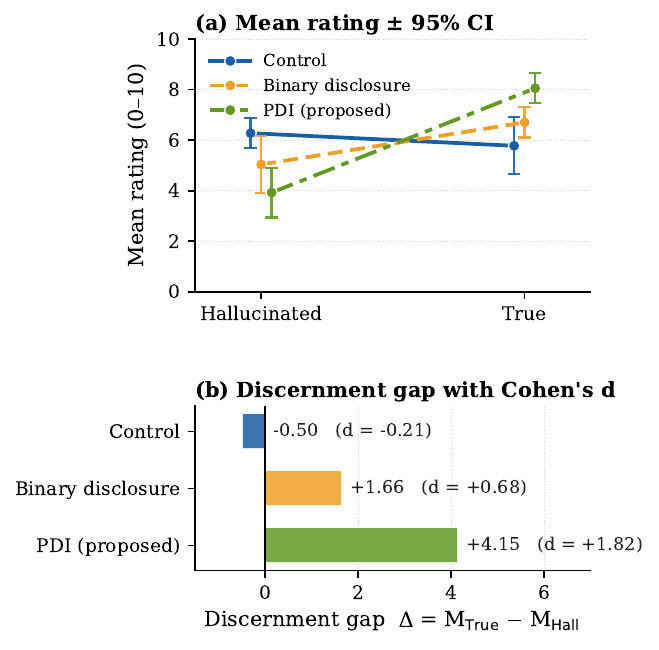}
    \caption{User-study results ($N=243$ ratings: 81 participants $\times$ 3 ratings each). \textbf{(a)} Mean rating $\pm$ 95\% CI shows the Interface~$\times$~Veracity interaction ($\chi^{2}(2)=31.32$, $p<.001$). \textbf{(b)} Discernment gap $\Delta = M_{\mathrm{True}} - M_{\mathrm{Hall}}$ with Cohen's $d$ annotated; PDI produces the largest separation ($d=1.82$).}
    \label{fig:results}
\end{figure}

\paragraph{Stigmatization vs. Calibration (Binary vs. PDI):}
While both interventions reduced trust in hallucinations, they did so via distinct cognitive mechanisms (Figure \ref{fig:results}): \textbf{Binary Disclosure (Stigmatization):} The ``Made with AI'' label functioned as a crude penalty. It reduced trust in hallucinations ($M = 5.04$) but failed to meaningfully restore confidence in the truth ($M = 6.70$). \textbf{Provenance Density (Calibration):} In contrast, PDI produced a large separation between truth and fabrication ratings. It penalized ungrounded content nearly twice as severely as binary labels ($M=3.93$), while simultaneously elevating trust in verified content to the highest observed level ($M = 8.08$). Post-hoc Tukey HSD comparisons across the Interface~$\times$~Veracity estimated marginal means confirm that PDI significantly restores confidence in verified content compared to standard Binary Disclosure ($p = .039$), effectively reversing the Transparency Penalty.

\begin{table}[h]
\centering
\resizebox{\columnwidth}{!}{%
\begin{tabular}{lrrrc}
\toprule
\textbf{Interface} & \textbf{Rating (Hall)} & \textbf{Rating (True)} & \textbf{Gap ($\Delta$)} & \textbf{Effect Size ($d$)} \\
\midrule
Control & \textbf{6.28} & 5.78 & -0.50 & -0.21 \\
Binary & 5.04 & 6.70 & +1.66 & 0.68 \\
PDI & \textbf{3.93} & \textbf{8.08} & \textbf{+4.15} & \textbf{1.82} \\
\bottomrule
\end{tabular}%
}
\caption{Quantifying Discernment. PDI creates the largest separation ($d=1.82$) between truth and fabrication.}
\label{tab:results_summary}
\end{table}

Taken together, these results show that PDI improved discernment by raising confidence in grounded content while reducing confidence in hallucinations, whereas Binary Disclosure produced a smaller separation.

\section{Discussion}
\label{sec:discussion}

Our quantitative results confirm that Provenance Density indicators (PDI) significantly outperform both Control and Binary Disclosure conditions in restoring truth discernment. By triangulating these statistical findings with the technical audit ($N=200$) and qualitative participant feedback ($N=81$), we infer the cognitive mechanisms driving these effects.

\paragraph{Mechanism of the Fluency Trap: Heuristic Substitution.} The failure of participants to distinguish between truth and hallucination in the Control condition ($p=.43$) is consistent with a related form of heuristic substitution: in the absence of provenance signals, participants may have substituted \textit{veracity} (is this true?) with \textit{fluency} (does this sound professional?). This pattern parallels research on receptivity to superficially impressive but semantically vacuous statements \cite{pennycook2015reception}, although our stimuli presented meaningful but false factual claims rather than vacuous prose.

Multiple participants explicitly described this strategy. P25 noted, ``They sound believable and well written... It didn't seem made up to me,'' while P81 judged accuracy based on ``how the text was worded and flowed.'' These comments support the interpretation that when the cost of generating professional-sounding text approaches zero, style ceases to be a reliable proxy for substance.

\paragraph{The ``Warning'' Effect of Binary Labels.} Prior work shows context-dependent effects of AI labels. In experiments with news headlines, AI labels reduced perceived accuracy even for true or human-generated headlines \cite{altay2024skeptical}; labels also reduced belief in misleading AI-generated image posts \cite{wittenberg2025labeling}, whereas a health-content experiment found no significant overall label effect and only nonsignificant reductions for accurate content \cite{li2024impact}. Our analysis suggests that Binary Disclosure operated through a blunt mechanism of \textit{epistemic stigmatization}: users interpreted the label not as a transparency aid, but as a risk marker.

P23 described the interaction vividly: ``The AI banner seemed like a warning vs being informative.'' In our study, this description supports the interpretation that the binary label operated as a risk cue rather than a verification aid. This context-specific penalty also aligns with concerns about the normative fairness of mandatory disclosure \cite{hosseini2025disclosing}, as it punishes the \textit{use} of the tool rather than the \textit{accuracy} of the content \cite{cheong2025penalizing}.

\paragraph{Calibrated Reliance and Cognitive Offloading.} In the PDI condition, participant strategies shifted from intrinsic text evaluation to extrinsic evidence evaluation, with users treating the Provenance Density score as an evidence cue. A critical question in HAI is whether such ``cognitive offloading'' functions as extended cognition \cite{chirayath2025cognitive,clark1998extended}. Our technical validation (Section \ref{sec:validation}) suggests cautious support, but with important limits. The metric's \textbf{Consistency Veto}—demonstrated by the suppression of scores in ambiguous queries such as the Indonesia capital transition—provides a partial safety rail for this offloading, not a guarantee. Users like P6 (``One of the passages also had a measurement of claims verified... I was more inclined to believe that'') are not blindly trusting the AI; they are trusting a \textit{verified attribute} of the information. However, because high-density misconceptions can still evade the veto, the interface should be understood as improving truth discernment rather than eliminating the ``False Assurance'' pitfall common in retrieval systems.

\paragraph{Ecological Alignment:} The metric's sensitivity to information age (Ecological Sensitivity) aligns user trust with the stability of knowledge. As shown in our technical results, established facts yielded higher density signals ($M=0.79$) than emerging dynamic topics ($M=0.64$). PDI therefore functions as a ``Time-Aware'' signal, implicitly guiding users to be more skeptical of breaking news (FreshQA) than settled science (TruthfulQA), mirroring the actual epistemic reliability of those domains.

\section{Limitations and Future Work}
\label{sec:limitations}

\paragraph{From Oracle to Deployment.} The Oracle protocol in Section~\ref{sec:oracle} shows that PDI has interaction value when the signal is correct, but deployment depends on how closely the live pipeline approaches that ideal. Retrieval can reinforce popular misconceptions, while the cross-sample consistency gate cannot detect false answers that recur consistently. The user-study result should therefore be interpreted as an upper bound on interface efficacy under correct signaling; a key next step is to study near-miss cases, especially false-positive high-density signals. A second design caveat is that the $3\times3$ Latin-square assignment did not fully cross interface, veracity, and topic: the PDI--Hallucinated cell was measured on Matcha, whereas the Control--Hallucinated cell averaged over Silk and Matcha. Thus, the observed PDI penalty ($M=3.93$) should be interpreted as strong evidence for the interface under these stimuli, rather than as a fully topic-independent estimate; a fully crossed replication is a clear next step.

\paragraph{Ecological Conservatism (The Cold Start Problem). }Provenance Density is inherently conservative: it privileges established consensus ($M=0.79$) over emerging novelty ($M=0.64$). Legitimate new information, such as breaking news or novel scientific hypotheses, may lack the citation network required to achieve high density ($S_{c} = \emptyset$), risking classification of ``the new'' as ``the unverified.'' Future iterations should integrate temporal metadata to distinguish low provenance due to fabrication from low provenance due to novelty.

\paragraph{Latency Constraints.}
The ``Consistency Veto'' is computationally costly: with audit-measured inference latencies ranging from 17.0s (Dynamic) to 26.4s (Static), $D(T)$ is unsuitable for synchronous, turn-by-turn chat. As noted in Section~\ref{sec:validation}, the higher Static latency reflects longer TruthfulQA generations rather than a swapped label. Future work should explore asynchronous interface patterns, such as ``Check in Background'' or ``Deep Check,'' that integrate verification latency without blocking interaction. Because latency was not experimentally manipulated in our study, future experiments should also test whether this friction functions as a costly signal that improves discernment, or merely as an engineering constraint.

\paragraph{Adversarial Gaming and Automation Bias.}
PDI also introduces second-order risks. The cubic MatchRatio term in Section~\ref{sec:metric} is designed to reduce citation-laundering by suppressing weak surface overlap, but the technical audit also surfaces vulnerabilities including domain spoofing, empty-keyword exploitation, SEO-style inflation, and mimetic misconception inheritance from open-web consensus. Hardening the reference implementation will require stricter domain parsing, conservative handling of vague claims, caps on per-domain contribution, and checks for duplicated evidence.
The risk of \textit{automation bias} \cite{cummings2017automation} remains: if users transition from ``reading the text'' to ``scanning for the green bar,'' future designs should add frictional elements that periodically force manual verification.

\section{Conclusion}
\label{sec:conclusion}

This paper argues that AI-generated misinformation is a problem of \textit{signaling} as well as \textit{generation}. As fluency becomes cheap, style no longer reliably signals substance: without external scaffolding, users in our study could not distinguish grounded truth from high-fluency hallucination.

Binary ``Made with AI'' labels address this crisis through identity disclosure, but our findings show that they act as crude warnings, producing a Transparency Penalty in which accurate AI-generated information is discounted alongside fabrication. Provenance Density offers a more calibrated alternative by visualizing verification rather than authorship. In the user study, PDI created a large positive discernment gap (+4.15), restoring users' ability to separate truth from fabrication under correct signaling. The technical audit qualifies this promise by showing that high-density evidence can still support misconceptions.

AI transparency should therefore shift from detecting \textit{who} wrote a text toward visualizing \textit{what} supports it, with interfaces that make verification visible, intuitive, and contestable.

\section*{Ethical Statement}

This research was conducted in accordance with the ethical
standards of the University of Tokyo institutional review process. We recruited 81 participants for the user study via
Prolific, all of whom provided informed consent before engaging
with the study materials. Prolific submission IDs were collected
solely for payment processing, stored separately from response
data, and are not retained in the released datasets; no other
personally identifying information was collected. All analyses
and shared materials use pseudonymous participant codes
(e.g., \texttt{P\_G1\_01}); raw survey exports are excluded
from the open-science release in accordance with the consent
form, while the verbatim consent protocol and the de-identified
participant ratings are included. Participants were compensated
for their time in accordance with fair labor standards.

Beyond procedural compliance, we acknowledge the broader sociotechnical implications of proposing Provenance Density as a credibility standard. While PDI effectively mitigates hallucinations in digitized high-resource domains, it introduces a risk of \textit{Systemic Authority Bias} \cite{milgram1963behavioral,jackson2012people}. By algorithmically privileging information with a traceable digital lineage, citation-based metrics may inadvertently disenfranchise non-digitized knowledge systems—including indigenous epistemologies, oral histories, and low-resource languages—that lack dense citation networks. There is a specific risk that users may conflate ``unverified'' (lack of digital evidence) with ``untrue'' (falsehood). Consequently, we advocate that PDI be deployed strictly as a verification tool for digitized consensus, rather than a universal arbiter of truth, to prevent the exacerbation of existing epistemic exclusions.

\section*{Acknowledgments}

This work was supported in part by JSPS KAKENHI Grant Numbers 25KK0001 (Fund for the Promotion of Joint International Research (Fostering Joint International Research)) and 25K21241 (Grant-in-Aid for Early-Career Scientists), and by JST Moonshot R\&D Grant JPMJMS2012.

\bibliographystyle{named}
\bibliography{ijcai26}


\end{document}